\documentclass[letterpaper, 10 pt, conference]{ieeeconf}  

\IEEEoverridecommandlockouts                              

\usepackage{graphics} 
\usepackage{epsfig} 
\usepackage{mathptmx} 
\usepackage{times} 
\usepackage{amsmath} 
\usepackage{amssymb}  

\usepackage{dirtree} 
\usepackage{siunitx}
\usepackage{multirow}
 \usepackage{booktabs} 
\usepackage{tikz}

\usepackage{amssymb}
\usepackage{pifont}
\newcommand{\cmark}{\ding{51}}%
\newcommand{\xmark}{\ding{55}}%

\definecolor{SS-background}{HTML}{000000}
\definecolor{SS-car}{HTML}{6DA3FD}
\definecolor{SS-spray}{HTML}{E60000}
\usepackage{bm} 

\usepackage{graphicx}
\usepackage{url}

\title{\LARGE \bf
SemanticSpray++: A Multimodal Dataset for Autonomous Driving in Wet Surface Conditions
}

\author{Aldi Piroli$^{1}$, Vinzenz Dallabetta$^{2}$, Johannes Kopp$^{1}$, Marc Walessa$^{2}$, \\ Daniel Meissner$^{2}$, and Klaus Dietmayer$^{1}$
\thanks{$^{1}$ Institute of Measurement, Control, and Microtechnology, Ulm University, Germany {\tt\small \{firstname.lastname\}@uni-ulm.de}}
\thanks{$^{2}$ BMW~AG, Petuelring 130, 80809~Munich,~Germany {\tt\small \{vinzenz.dallabetta, marc.walessa\}@bmw.de} and {\tt\small daniel.da.meissner@bmwgroup.com}}%
}

\newcommand\copyrighttext{%
	\footnotesize \copyright\,2024 IEEE. Personal use of this material is permitted. Permission from IEEE must be obtained for all other uses, in any current or future media, including reprinting/republishing this material for advertising or promotional purposes, creating new collective works, for resale or redistribution to servers or lists, or reuse of any copyrighted component of this work in other works.}%
\newcommand\copyrightnotice{%
	\begin{tikzpicture}[remember picture,overlay]%
	\node[anchor=south,yshift=10pt] at (current page.south) {\fbox{\parbox{\dimexpr\textwidth-2cm}{\copyrighttext}}};%
	\end{tikzpicture}%
	\vspace{-10pt}%
}

\begin{document}

\maketitle
\copyrightnotice
\thispagestyle{empty}
\pagestyle{empty}

\begin{abstract}
Autonomous vehicles rely on camera, LiDAR, and radar sensors to navigate the environment. 
Adverse weather conditions like snow, rain, and fog are known to be problematic for both camera and LiDAR-based perception systems.  
Currently, it is difficult to evaluate the performance of these methods due to the lack of publicly available datasets containing multimodal labeled data. 
To address this limitation, we propose the SemanticSpray++ dataset, which provides labels for camera, LiDAR, and radar data of highway-like scenarios in wet surface conditions.
In particular, we provide 2D bounding boxes for the camera image, 3D bounding boxes for the LiDAR point cloud, and semantic labels for the radar targets.
By labeling all three sensor modalities, the SemanticSpray++ dataset offers a comprehensive test bed for analyzing the performance of different perception methods when vehicles travel on wet surface conditions. 
Together with comprehensive label statistics, we also evaluate multiple baseline methods across different tasks and analyze their performances. 
The dataset will be available at \url{https://semantic-spray-dataset.github.io}
\end{abstract}
\section{Introduction}
The pursuit of achieving autonomous driving has accelerated research across various disciplines. 
Notably, advancements in computer vision applications for diverse sensor modalities such as camera, LiDAR, and radar have significantly benefited from this endeavor. 
Consequently, there has been an unparalleled advancement in tasks like object detection and semantic segmentation.
One of the main factors for innovations in these fields has been the advancements in deep learning methods. 
These advances have been possible mainly as a result of increases in computing power and data availability. 
\begin{figure}[t!]
    \centering
        \includegraphics[width=\columnwidth]{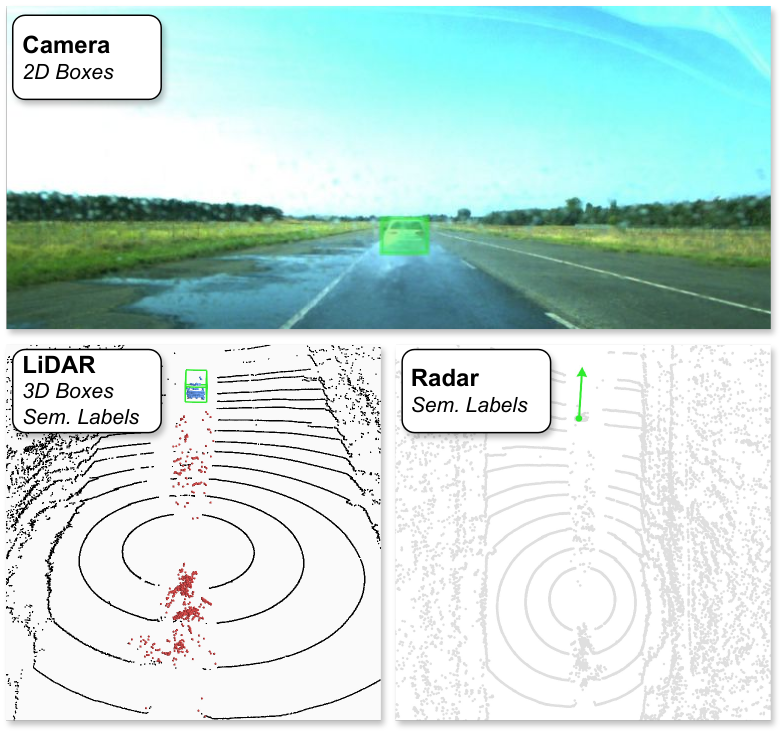}
    \caption{The proposed SemanticSpray++ dataset offers multimodal labels across camera, LiDAR, and radar sensors for testing the effect of spray on perception systems.
    \textbf{Top}: shows the camera image with overlayed 2D ground truth bounding box (in green) of the vehicle in front.
    \textbf{Bottom-left}: shows the captured LiDAR scan, where the 3D ground truth bounding box (in green) represents the leading vehicle. 
    Additionally, each point has an associated semantic label, where the colors represent $\color{SS-background}{\bullet}~$\textit{background}
$\color{SS-car}{\bullet}~$\textit{foreground}
$\color{SS-spray}{\bullet}~$\textit{noise}.
\textbf{Bottom-right}: shows the radar target represented by the Doppler velocity vector (green arrow). 
We also overlay the LiDAR scan for visualization purposes in gray.
    }
    \label{Fig:teaser}
\end{figure}

Despite the many improvements, the task of autonomous driving is not yet considered complete. 
One of the many reasons is that neural networks perform unexpectedly when tested in a different domain than the one used during training~\cite{du2022vos, piroli2022detection, guo2017calibration, piroli2023ls, devries2018learning}. 
For example, modern object detectors tend to detect unknown objects as one of the training classes, often with high confidence scores (e.g., an animal is classified as a pedestrian)~\cite{du2022vos}.
A more mundane but far more common example is the performance degradation of camera and LiDAR-based sensor systems in adverse weather conditions such as rain, snow, and fog~\cite{piroli2022robust, dreissig2023surveyAdverseWeather, piroli2023SemanticSpray, piroli2023towards}. 
For example, in foggy and rainy conditions, the camera's view is greatly reduced, resulting in fewer objects being detected.  
In LiDAR sensors, the water particles that make up rain, spray, and snow can cause the measurement signal to be scattered, resulting in missed point detections.
In addition, these same water particles can cause partial or total reflection of the signal, resulting in additional unwanted noise in the measurements. 
These effects can cause detectors that perform well in good weather conditions to missdetect objects and introduce false positive detections in the perceived environment.
Since autonomous vehicles rely heavily on these sensors, such unexpected behavior can have very serious consequences and, in extreme cases, pose a threat to passengers and other road users. 

Therefore, it is important to test perception methods in different weather conditions thoroughly. 
However, few datasets are currently available for testing such systems, and even fewer provide labeled data for all common sensor modalities (i.e., camera, LiDAR, and radar).
For example, the Waymo Open dataset~\cite{sun2020waymo} contains a large number of scenes in rainy conditions and provides labels for both camera and LiDAR sensors. 
However, no radar data is available.
In addition, general-purpose datasets like the Waymo Open or nuScenes datasets~\cite{caesar2020nuscenes}  lack the systematic testing of a specific weather condition and all of its many variations. 
For example, on wet surfaces, the resulting spray effect is highly dependent on driving speed and vehicle type~\cite{shih2022reconstruction}.

To address some of the problems mentioned above, we propose the SemanticSpray++ dataset, which has labels for vehicles traveling in wet surface conditions at different speeds.
We provide labels for camera, LiDAR, and radar sensors for object detection as well as for semantic segmentation tasks.
Our work is based on the RoadSpray dataset~\cite{linnhoff2022roadspray}, which contains the raw and unlabeled recordings, and in our previous publication, where we published the SemanticSpray dataset~\cite{piroli2023SemanticSpray} that provides semantic labels for the LiDAR point clouds.
In this paper, we extend our previous work by additionally labeling 2D bounding boxes for the camera image, 3D bounding boxes for the LiDAR point clouds, and semantic labels for the radar targets.
An example of the different annotated modalities is shown in Fig.~\ref{Fig:teaser}.
As the data extensively covers different speeds, vehicles, and amounts of surface water, it provides a unique test bed where 2D and 3D object detectors and semantic segmentation methods can be tested to understand their limitations in this particular weather effect.
In addition, we test different baseline perception methods like 2D and 3D object detectors and 3D semantic segmentation networks and analyze the effect that spray has on their performance.

In summary, our main contributions are:
\begin{itemize}
        \item We extend the SemanticSpray dataset to include multimodal labels for vehicles traveling in wet surface conditions. 
        \item We provide 2D bounding box labels for the camera images, 3D bounding boxes for the LiDAR point clouds, and semantic labels for the radar targets.
        \item We provide label statistics on both object-level and point-wise semantic.
        \item We test popular perception methods across different tasks and analyze how their performance is affected by spray.
\end{itemize}

\section{Related Work}
\subsection{Datasets for Autonomous Driving}
The recent advances in the methods for autonomous driving have been made possible in part by the influx of large and diverse datasets.
The KITTI dataset~\cite{geiger2013KITTI} pioneered this field by proposing annotated labels for both LiDAR and camera images, allowing 2D and 3D object detectors to be tested.
The SemanticKITTI dataset~\cite{behley2019semantickitti} provides additional LiDAR point-wise semantic labels, allowing training and testing of semantic segmentation networks.  
The nuScenes dataset~\cite{caesar2020nuscenes} is a popular dataset that provides labels for many tasks, among which are multi-camera object detection, semantic segmentation, and 3D object detection.
It is recorded in urban scenarios with different weather conditions in North America and Southeast Asia.
The Waymo Open dataset~\cite{sun2020waymo} is a large-scale dataset that provides annotated camera and LiDAR point clouds in both urban and extra-urban scenarios in sunny and rainy conditions. 
Additional datasets such as Argoverse~\cite{Argoverse} and ZOD~\cite{alibeigi2023ZOD} also provide a large and diverse set of annotated frames for autonomous driving applications.
Furthermore, there are many datasets which only provide single modalities (e.g., camera only) data annotations like Cityscapes~\cite{cordts2016cityscapes} and BDD100K~\cite{yu2020bdd100k}.

\subsection{Adverse Weather Datasets for Autonomous Driving}
Recently, a few datasets have been proposed for autonomous driving applications where the focus is on adverse weather conditions~\cite{dreissig2023surveyAdverseWeather}. 
The Seeing Through Fog dataset (STF)~\cite{bijelic2020STF} provides annotated open-world recordings while driving in foggy conditions for both camera and LiDAR point clouds. 
SemanticSTF~\cite{xiao20233dSemanticSTF} has recently extended the STF dataset to provide semantic labels.
The DENSE dataset~\cite{heinzler2020cnn} is recorded in a weather chamber where artificial fog, snow, and rain are measured with a LiDAR sensor with a simulated urban scenario in the background. 
The ADUULM dataset~\cite{pfeuffer2020aduulm} provides semantic labels for camera and LiDAR point clouds in diverse weather conditions. 
The WADS dataset~\cite{kurup2021WADS} focuses on snowy conditions, providing semantic segmentation labels for the 3D LiDAR point clouds. 
The CADC dataset~\cite{pitropov2021CADAC} instead contains 3D bounding boxes for LiDAR point clouds in snowy conditions. 
The RADIATE dataset~\cite{sheeny2021radiate} also focuses on adverse weather and is one of the few datasets that provides object-level annotations for the radar sensor. 
The RoadSpray dataset~\cite{linnhoff2022roadspray} provides an extensive list of recordings in wet surface conditions.
It contains scenes in a highway-like environment at different speeds, with camera, LiDAR, and radar measurements. 
However, the dataset only contains raw recordings without any annotations for any of the sensor modalities. 
The SemanticSpray dataset~\cite{piroli2023SemanticSpray}, which is based on a subset of scenes of the RoadSpray dataset, provides semantic segmentation labels for the LiDAR scenes, differentiating between the foreground objects in the scenes,  background points, and noise points like spray and other adverse weather artifacts. 
The proposed SemanticSpray++ dataset is based on the RoadSpray and SemanticSpray datasets and aims to extend the label annotations to different sensor modalities and formats. 

\section{SemanticSpray++ Dataset}
This section introduces the SemanticSpray++ dataset and gives an overview of the scenarios, the recording setup, the data annotation, the label format, and label statistics. 
\begin{figure}[t!]
    \centering
        \includegraphics[width=\columnwidth]{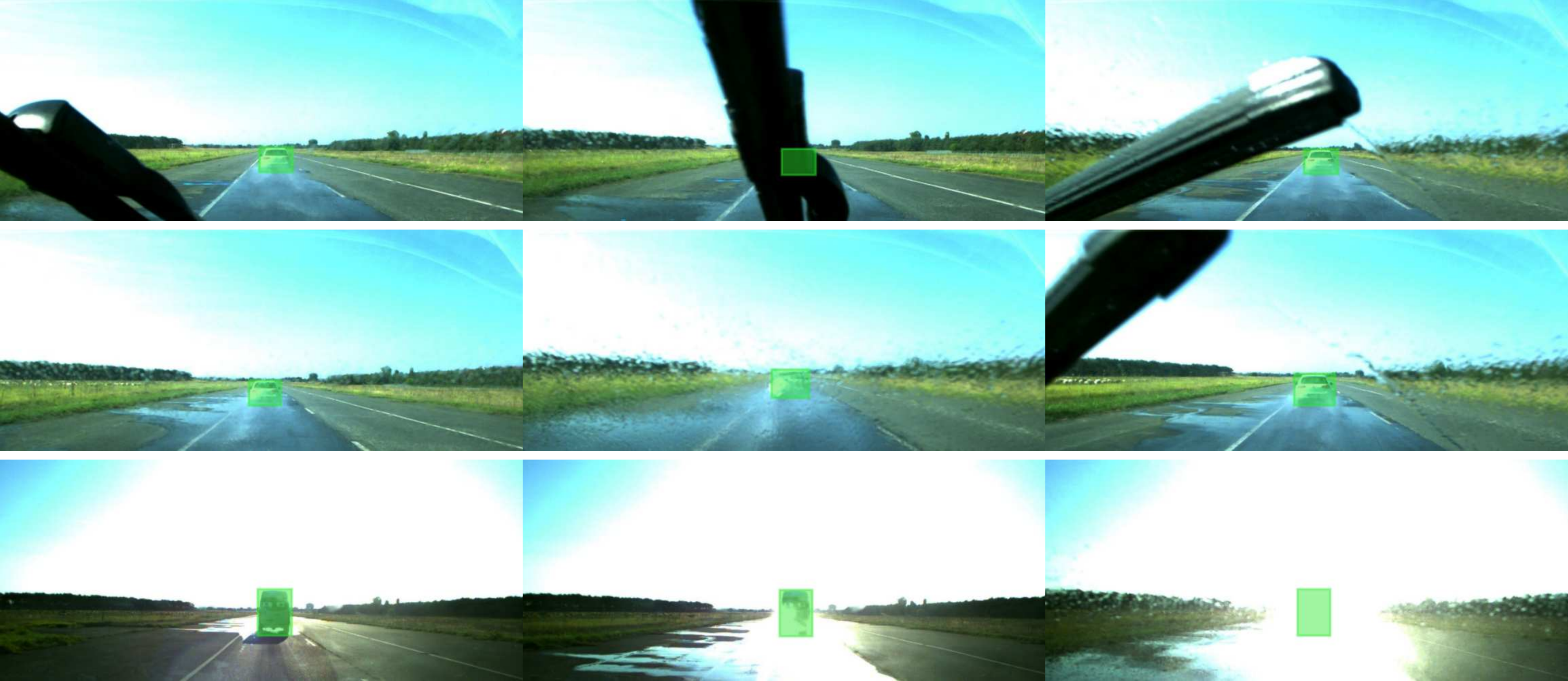}
    \caption{    
Overview of some of the scenes present in the proposed dataset.
\textbf{Top row}: show the occlusion effect caused by the windshield wipers. 
\textbf{Middle row}: shows the blurriness effect caused by the spray particles generated by the leading vehicle.
\textbf{Bottom row}: shows how sunlight directly reflecting off the camera sensor or on the wet surface leads to locally overexposed images, which block the leading vehicle from the field of view. 
We show with green boxes the provided 2D box annotations.
    }
    \label{Fig:camera_label_problems}
\end{figure}
\subsection{Scenario Setup}
The SemanticSpray++ dataset provides LiDAR, camera, and radar labels for a subsection of the scenes of the RoadSpray dataset~\cite{linnhoff2022roadspray}.
The RoadSpray dataset itself provides unlabeled data for vehicles traveling on wet surfaces at different speeds in a highway-like scenario. 
In the relevant experiments, the ego vehicle follows a leading vehicle at a fixed distance while traveling at different speeds.
The distances between the ego and the leading vehicles are between \SI[per-mode=symbol]{20}{\m} and \SI[per-mode=symbol]{30}{\m}, whereas the traveling speeds are in the range \SIrange[range-phrase=--,range-units=single, per-mode=symbol]{50}{130}{\km\per\hour}, with \SI[per-mode=symbol]{10}{\km\per\hour} increments. 
There are two types of lead vehicles: a small size car and a large van vehicle.
The experiments are conducted in an empty airstrip field to recreate a highway-like scenario.
The ego vehicle is equipped with the following sensors:
\begin{itemize}
    \item High-resolution LiDAR (top-mounted),
    \item Long-range radar (front-mounted),
    \item Camera sensor (front-mounted).
\end{itemize}
For more information on the raw data recording and a more detailed sensor description, we refer the reader to the original dataset publication~\cite{linnhoff2022roadspray}.

\subsection{Data Format and Annotation}
As the RoadSpray dataset provides raw unlabeled data in a \texttt{rosbag} format, we extract the camera, LiDAR, and radar data.
Because the different sensors record at different frequencies, we use the LiDAR sensor as the synchronization signal in the extraction process.    
The LiDAR point clouds are saved in binary files with each point having features $(x,y,z,\text{intensity}, \text{ring})$, where $(x,y,z)$ is the 3D position of the points, intensity is a value ranging from $0-255$ which quantifies the calibrated-intensity of the point, and ring represents which of the LiDAR layers the point originated from.   
The radar sensor data is also saved in a binary file, where each point has features $(x,y,v_x, v_y)$, where $(x,y)$ is the 2D Cartesian position of the point and $(v_x, v_y)$ are the components of the Doppler velocity vector. 
The camera sensor data is saved in an RGB \texttt{.jpg} image file of size $2048\times1088$ pixels.

The SemanticSpray dataset~\cite{piroli2023SemanticSpray} provides semantic labels for the LiDAR point clouds, assigning to each point one of the three possible labels: \textit{background} (vegetation, building, road, \ldots),  
 \textit{foreground} (leading vehicle),   and \textit{noise} (spray and other weather artifacts). 
The labels are provided in a binary file with label mapping $\{0: \textit{background}, 1: \textit{foreground}, 2: \textit{noise}\}$.

\begin{figure}[t!]
    \centering
        \includegraphics[width=\columnwidth]{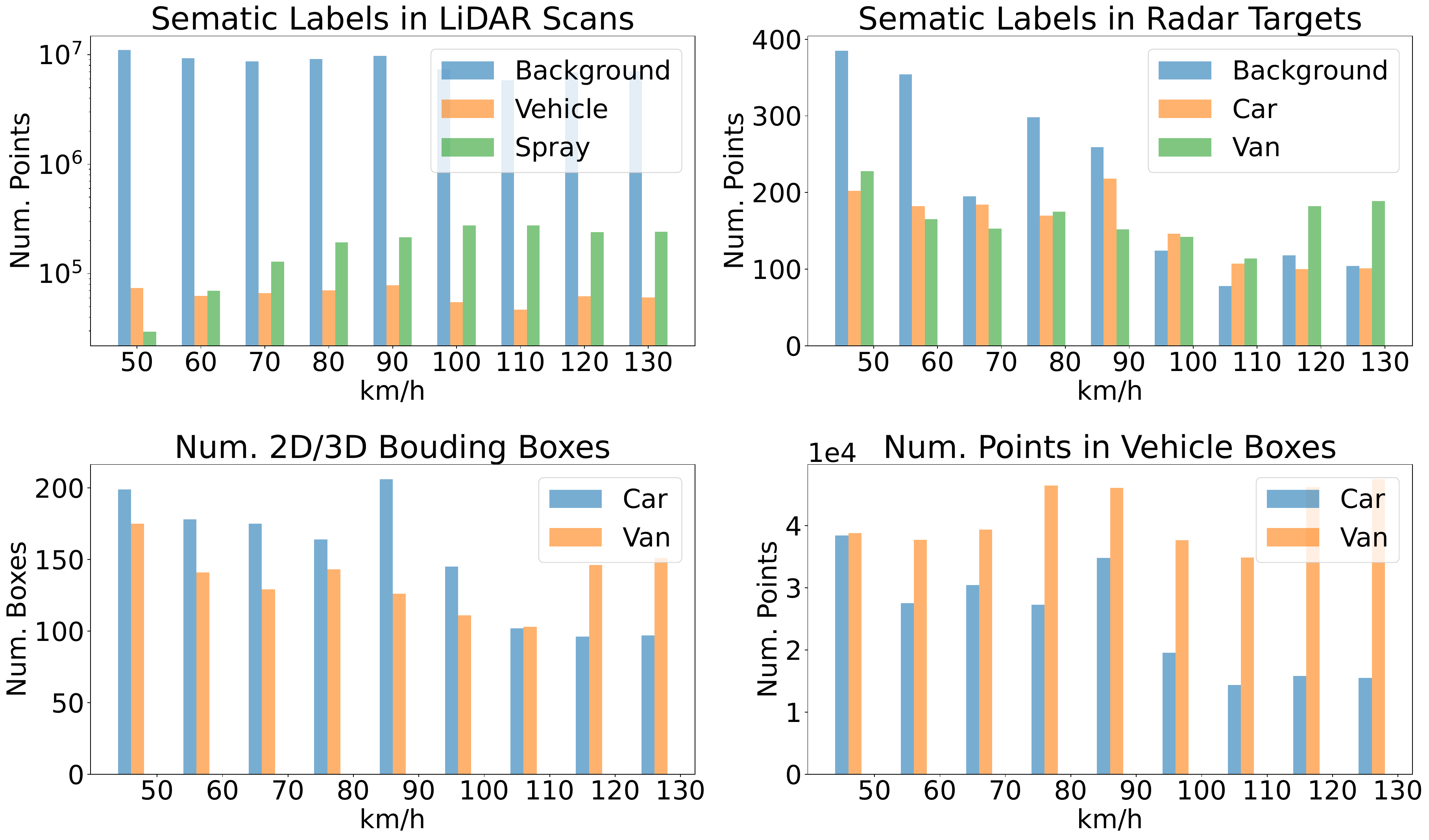}
    \caption{    
    Statistics of the proposed dataset.
    \textbf{Top-left}: shows the distributions of the LiDAR-based semantic labels among different velocities.   
    \textbf{Top-right}: shows the distributions of the radar-based semantic labels among different velocities.
    \textbf{Bottom-left}: shows the number of 2D and 3D object box annotations in the camera and LiDAR point clouds. 
    The number of boxes matches among the different modalities as both sensors always capture the leading vehicle.
    \textbf{Bottom-right}: shows the number of vehicle points in the 3D LiDAR bounding boxes at different speeds.
    }
    \label{Fig:dataset_statistics}
\end{figure}

With the SemanticSpray++ dataset, we extend the available labels to 2D bounding boxes for the camera image, 3D bounding boxes for the LiDAR point cloud, and semantic labels for the radar targets.
We select a subset of $36$ scenes from the SemanticSpray dataset, choosing scenes with different traveling speeds and vehicle distances. 
In Fig.~\ref{Fig:teaser} and Fig.~\ref{Fig:camera_label_problems}, we show examples of all annotation types, and in Fig.~\ref{Fig:dataset_statistics}, an overview of the dataset statistics.

\textbf{LiDAR 3D Bounding Box Annotation.}
We annotate the 3D LiDAR point clouds using bounding boxes with the format $[x,y,z,w,h,l,\theta]$, where $(x,y,z)$ is the center of the bounding box, $(w,h,l)$ are its height, width and length and $\theta$ the orientation around the $z$-axis.
Since there are two different types of leading vehicles, we label them as separate classes, namely \textit{Car} and \textit{Van}.
We store the labels in \texttt{.json} files, which are easy to parse and still human-readable.

\textbf{Camera 2D Bounding Box Annotation.}
For the camera data, we annotate each image by assigning a bounding box to each object in the scene.
We use the format [\textit{top-left}, \textit{top-right}, \textit{bottom-left}, \textit{bottom-right}], where for each point, we give the $(x,y)$ coordinates in the camera image. 
For each box, we use the same class categories \textit{Car} and \textit{Van} as for the LiDAR data.
In the recorded data, there are many instances where the leading vehicle is totally occluded or only partially visible in the camera image. 
This is mainly due to windshield wipers blocking the field of view or water particles caused by the vehicle in front blocking or blurring the camera view. 
In addition, there are many cases where the camera image is locally overexposed due to sunlight reflecting directly onto the camera sensor or off the wet surface.   
To address this issue during the labeling process, we interpolate the occluded bounding boxes between two visible camera frames. 
We report some examples of the effects and the interpolation process in Fig.~\ref{Fig:camera_label_problems}.

\textbf{Semantic Labels of Radar Targets.}
Since the radar and LiDAR point clouds are calibrated, we use a semi-automatic approach for the semantic labeling of the radar targets. 
First, we project the radar points in the LiDAR point cloud coordinate frame. 
Then, using the 3D bounding boxes described in the previous section, we check which of the radar targets are contained within the LiDAR-based boxes and automatically label them as either \textit{Car} or \textit{Van} (depending on the leading vehicle type) or as \textit{Background} if they do not belong to any of the vehicles in the scene. 
After the automatic labeling, we manually check each radar point cloud and fix possible incorrect labels.


\subsection{Dataset Toolkit}
Together with the dataset, we provide a toolkit that provides useful data processing and visualization scripts. 
Among these, we provide a PyTorch data loader for the 3D object detection framework OpenPCDet~\cite{openpcdet2020}, which allows for easy testing of different object detectors and for the SPVCNN framework~\cite{tang2020searching}, which allows to train and test multiple semantic segmentation networks.
Additionally, we provide useful scripts to convert the labels into different formats (e.g., COCO, YOLO, OpenLABEL~\cite{hagedorn2021open}).

\section{Experiments}
In this section, we report the evaluation of baseline methods among different tasks. 
In particular, we aim to test how the performance of 3D LiDAR object detectors, 2D camera-based detectors, and 3D LiDAR semantic segmentation networks are affected when evaluating their performance in the adverse weather conditions present in SemanticSpray++.

\subsection{Experiment Setup}
\textbf{3D LiDAR Object Detector.}
As spray points heavily impact LiDAR sensors, we test the performance of three popular object detectors (PointPillars~\cite{lang2019pointpillars}, SECOND~\cite{yan2018second} and CenterPoint~\cite{yin2021center} with a pillar backbone) trained on the nuScenes dataset~\cite{caesar2020nuscenes}, using the respective OpenPCDet~\cite{openpcdet2020} implementation.
We follow the evaluation setup presented in~\cite{piroli2023towards}, where we first test the trained models directly on the SemanticSpray++ dataset.
In addition, we fine-tune the detectors on a small subset of scenes from SemanticSpray++ where no spray points are present. 
This allows us to reduce the domain gap caused by factors such as sensor placement and better understand the impact of spray on the detectors~\cite{piroli2023towards}. 
In contrast to the evaluation reported in~\cite{piroli2023towards}, we use the ground truth data provided in SemanticSpray++ instead of pseudo-labels. 
Moreover, we include scenes with both the \textit{Car} and \textit{Van} leading vehicle, instead of only using the \textit{Car} class.
For a detailed description of the fine-tuning process, we refer the reader to~\cite{piroli2023towards}.
As evaluation metrics for 3D object detection, we use the Average Precision (AP) metric with Intersection over Union (IoU) at $0.5$ and class-wise mean AP (mAP). 
We compute it at different ranges, namely  \SIrange[per-mode=symbol]{0}{25}{\m} and $>$ \SIrange[per-mode=symbol, range-units=single]{0}{25}{\m}.
As the nuScenes dataset provides multiple vehicle categories, we use the following output mapping  $\{\textit{Truck, Construction Vehicle, Bus, Trailer}\}\rightarrow \textit{Van}$ for the large-vehicle detections.

\textbf{2D Object Detection.}
For testing the performance of 2D camera-based detections, we use the popular YOLO~\cite{redmon2016you} object detector, adapting the implementation provided by~\cite{yolov8ultralytics}.
We use the YOLOv8m (mid-sized) model trained on the COCO dataset~\cite{lin2014microsoftCOCO}. 
Additionally, we train the same model using the default configurations on the Argoverse dataset~\cite{Argoverse}, which contains annotated images from the perspective of the ego-vehicle. 
As mentioned in the previous section, camera images are affected in different ways by wet surface conditions. 
To overcome some of these effects (i.e., partial or total occlusion, blur, and overexposure), we test the performance of the YOLOv8m models with two different object trackers (ByteTrack~\cite{zhang2022bytetrack} and BoT-SORT~\cite{aharon2022bot}), which allow the use of temporal information when detecting objects, even in the presence of occlusion.
As evaluation metrics, we use the standard AP at $0.5$ IoU.

\textbf{3D LiDAR Semantic Segmentation.}
We additionally test how the performance of semantic segmentation networks is affected when testing on scenes with spray.
For this purpose, we use SPVCNN~\cite{tang2020searching} as the base network and train on both the nuScenes~\cite{caesar2020nuscenes} and SemanticKITTI dataset~\cite{behley2019semantickitti} using the official implementations.
Since the classes of the training set do not all match the classes of the SemanticSpray++ dataset, a direct quantitative comparison of performance is not possible. 
Instead, we report the confusion matrix for each method, which provides insight into how the \textit{noise} points tend to be misclassified.

\subsection{Experiment Results}
\begin{figure*}[t!]
    \centering
        \includegraphics[width=\textwidth]{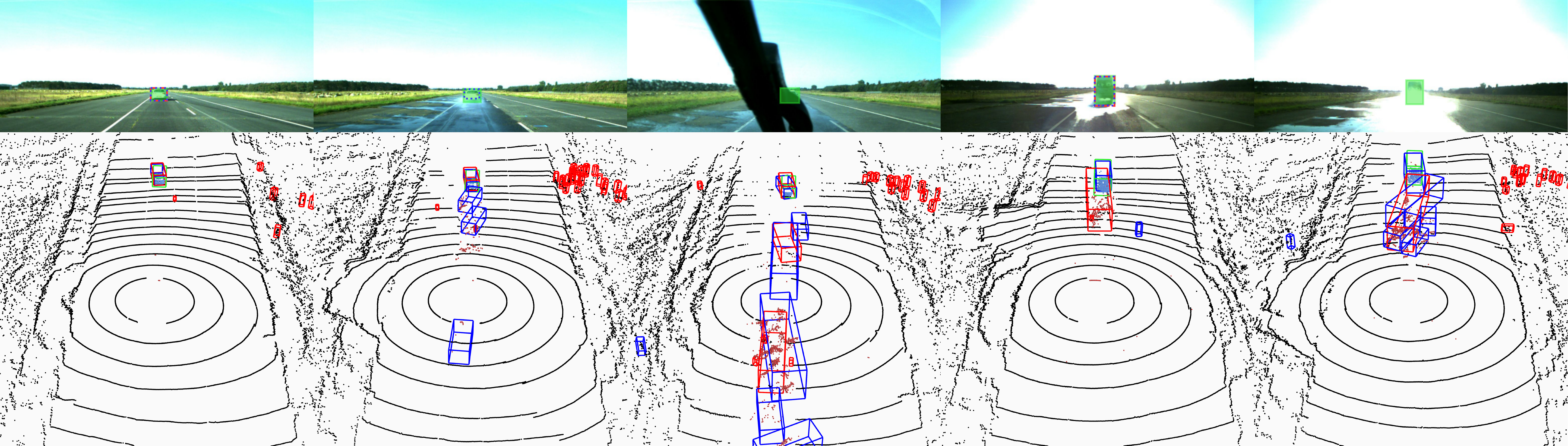}
    \caption{    
    Qualitative results for 2D and 3D object detectors tested on SemanticSpray++.
    \textbf{Top row}: shows the camera image with overlayed ground truth bounding boxes $\color{green}{\mathbf{-}}$, predictions using YOLOv8m $\color{red}{\mathbf{--}}$, and predictions using predictions using YOLOv8m + BoT-SORT $\color{blue}{\mathbf{--}}$.
    \textbf{Bottom row}: shows the LiDAR point could with ground truth boxes $\color{green}{\mathbf{-}}$, predictions from SECOND trained only on nuScenes $\color{red}{\mathbf{-}}$, and SECOND trained on nuScenes with additional fine-tuning on SemanticSpray++ $\color{blue}{\mathbf{-}}$.
    The semantic labels for the LiDAR point cloud have the following color map: $\color{SS-background}{\bullet}~$\textit{background}
$\color{SS-car}{\bullet}~$\textit{foreground}
$\color{SS-spray}{\bullet}~$\textit{noise}.
    }
    \label{Fig:qualitative}
\end{figure*}

\textbf{3D LiDAR Object Detector.}
\begin{table*}[t!]
    \centering
    \caption{Evaluation results of 3D object detectors trained on the nuScenes dataset only and fine-tuned on spray-free scans of the SemanticSpray++ dataset. 
    Results are in percentage.
    }
    \resizebox{1\textwidth}{!}{%

        \begin{tabular}{@{}clccclccclc@{}}
            \toprule
            \multirow{2}{*}{\textbf{Fine-Tuned}}    & \multirow{2}{*}{\textbf{Detector}} &
            \multicolumn{3}{c}{\textbf{\textit{Car} AP}} &
            \multicolumn{3}{c}{\textbf{\textit{Van} AP}}     &
            \multicolumn{3}{c}{\textbf{mAP}}                                                                                   \\
            \cmidrule(l){3-11}
                                                    &                                    &
            $0$-\SI{25}{\meter}                           & $>$\SI{25}{\meter}         & overall& $0$-\SI{25}{\meter} 
                                                    & $>$\SI{25}{\meter}         & overall& $0$-\SI{25}{\meter}  &
            $>$\SI{25}{\meter}              & overall                                                        \\



            \midrule
            \multirow{3}{*}{\xmark}                 & PointPillar~\cite{lang2019pointpillars}
                                                    &
            67.15                                   & 63.81
                                                    & 61.53
                                                    & 12.93
                                                    & 14.64
                                                    & 12.81
                                                    & 40.04                              & 39.22
                                                    & 37.17
            \\
                                                    & SECOND~\cite{yan2018second}
                                                    &
            \textbf{97.02}                          & \textbf{96.79}
                                                    & \textbf{96.07}
                                                    & 8.18
                                                    & \textbf{22.03}
                                                    & 9.53
                                                    & \textbf{52.60}                     &
            \textbf{59.41}                          & \textbf{52.80}
            \\
                                                    & CenterPoint~\cite{yin2021center}
                                                    &
            56.48                                   & 16.37
                                                    & 30.74
                                                    & \textbf{13.06}
                                                    & 10.86
                                                    & \textbf{13.06}
                                                    & 34.77                              & 13.62
                                                    & 21.90
            \\ \midrule
            \multirow{3}{*}{\cmark}                 & PointPillar~\cite{lang2019pointpillars}
                                                    &
            92.28                                   & 93.71
                                                    & 91.56
                                                    & \textbf{69.37}
                                                    & 73.90
                                                    & \textbf{72.31}
                                                    & \textbf{80.83}                     & 83.80
                                                    & \textbf{81.93}
            \\
                                                    & SECOND~\cite{yan2018second}
                                                    &
            \textbf{98.07}                          & \textbf{99.86}
                                                    & \textbf{97.64}
                                                    & 43.82
                                                    & 77.30
                                                    & 60.70
                                                    & 70.95                              &
            \textbf{88.58}                          & 79.17
            \\
                                                    & CenterPoint~\cite{yin2021center}
                                                    &
            92.77                                   & 89.58
                                                    & 88.77
                                                    & 44.08
                                                    & \textbf{85.28}
                                                    & 70.96
                                                    & 68.43                              & 87.43
                                                    & 79.86
            \\ \bottomrule
        \end{tabular}

    }
    \label{table:det3d}
\end{table*}
We begin our evaluation by testing the performance of  PointPillars, SECOND, and CenterPoint trained only on the nuScenes dataset and report the results in Table~\ref{table:det3d}.
We can observe that all detectors perform higher when detecting the \textit{Car} type. 
Additionally, we see that in most cases, the performance between $0$-\SI{25}{\meter} is higher than for $>$\SI{25}{\meter}.
This is expected as the density of LiDAR points diminishes with the distance from the sensor. 
Moreover, the total number of returned points is also reduced due to the scattering and partial/total occlusion effects. 
This can also be seen in Fig.~\ref{Fig:dataset_statistics}, where the number of points inside the bounding boxes decreases as the driving speed increases due to more spray.
When analyzing the results for the fine-tuned detectors, we see that all of the models benefit from it, greatly increasing the performance for both \textit{Car} and \textit{Van} classes.
When we compare the qualitative results from Fig.~\ref{Fig:qualitative}, we notice that both fine-tuned and non-fine-tuned detectors are affected by spray points, which cause false-positive detections to arise. 
However, we can observe that for the non-fine-tuned detectors, additional factors like the surrounding environment also cause false positive detections.

\textbf{2D Camera Object Detector.}
\begin{table}[t!]
    \centering
    \caption{Evaluation results on the camera-based 2D object detection task.
    The results show the YOLOv8m model trained on different datasets with and without object tracking as post-processing.
    Results are in percentage.
    }
    \resizebox{1\columnwidth}{!}{%

\begin{tabular}{@{}lcccc@{}}
\toprule
\textbf{Training Set}      & \textbf{Tracking}      & \textbf{\textit{Car} AP@0.5} & \textbf{\textit{Van AP@0.5}} & \textbf{mAP@0.5} \\ \midrule
\multirow{3}{*}{COCO~\cite{lin2014microsoft}}      & \xmark             & 75.05               & 73.60               & 74.33            \\
                           & ByteTrack~\cite{zhang2022bytetrack} & 80.61               & 72.36               & 76.49            \\
                           & BoT-SORT~\cite{aharon2022bot}   & \textbf{81.39}      & \textbf{74.17}      & \textbf{77.78}   \\ \midrule
\multirow{3}{*}{Argoverse~\cite{Argoverse}} & \xmark             & 75.20               & 73.61               & 74.41            \\
                           & ByteTrack~\cite{zhang2022bytetrack} & 82.65               & 75.03               & 78.84            \\
                           & BoT-SORT~\cite{aharon2022bot}   & \textbf{82.93}      & \textbf{75.96}      & \textbf{79.45}   \\ \bottomrule
\end{tabular}

    }
    \label{table:det2d}
\end{table}
In Table~\ref{table:det2d}, we report the results of the camera-based 2D object detector and the additional object tracking post-processing. 
We can see that the performance for both YOLOv8m trained on COCO and Argoverse are similar for both the \textit{Car} and \textit{Van} classes. 
We can also observe that object tracking substantially improves the performance of both models. 
For example, on YOLOv8m trained on Argoverse combined with BoT-SORT, the mAP improves by $+5.04\%$ points. 
Looking at the qualitative results of Fig.~\ref{Fig:qualitative}, we see that the performance of the object detector without any associated tracking is indeed affected by the occlusion of the windshield wipers, the blurriness of the water spray particles, and the locally overexposed images. 
The same figure shows that these problems are less pronounced when an object tracker is used as a post-processing step.

\textbf{3D LiDAR Semantic Segmentation.}
\begin{figure*}[t!]
    \centering
        \includegraphics[width=\textwidth]{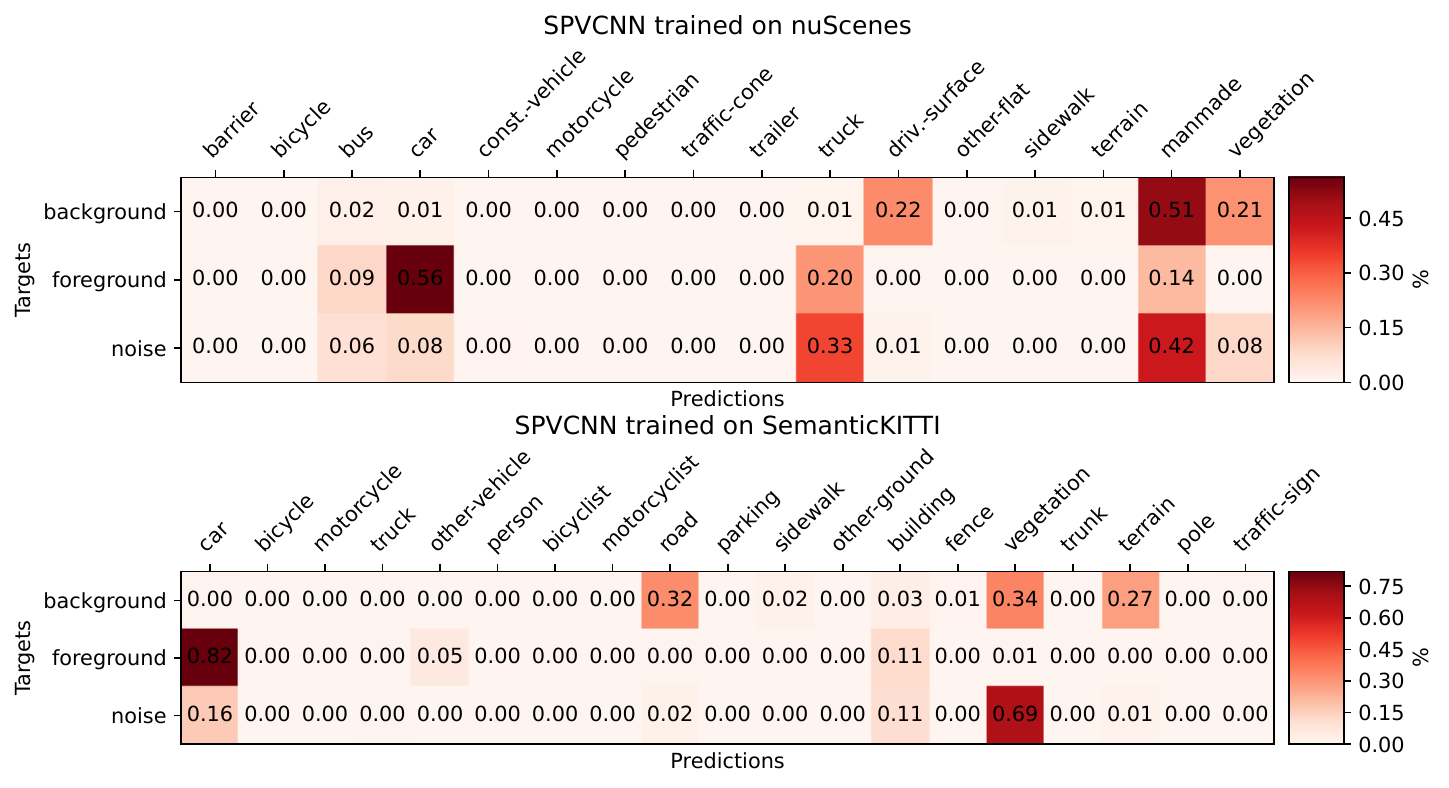}
    \caption{    
    Confusion matrix of SPVCNN trained on the nuScenes-semantic and SemanticKITTI datasets and evaluated on the SemanticSpray++ dataset. 
    Notice that, as the training labels do not match the test labels, the matrices are not square. 
    Additionally, small values ($<0.01$) are truncated to $0$ for visualization purposes.
    }
    \label{Fig:confusion_matrix}
\end{figure*}
As mentioned before, we provide the confusion matrices for SPVCNN trained on different semantic segmentation datasets and report it in Fig.~\ref{Fig:confusion_matrix}.
When observing SPVCNN trained on nuScenes, we see that the \textit{background} class is mainly classified by the network as \textit{man-made}, \textit{vegetation} and \textit{drivable-surface}.
Similar predictions can be observed when SPVCNN is trained on the SemanticKITTI dataset.
The foreground class, which consists of points belonging to two different lead vehicles, is for the most correctly associated with the vehicle classes of the two datasets. 
The noise class, which contains spray points and other weather artifacts, is instead associated with different semantic classes.
For example, when the model is trained on nuScenes, the \textit{mandmade} and \textit{truck} classes are the two most associated classes.
For SPVCNN trained on SemanticKITTI, it is instead \textit{vegetation}.
These results highlight the overconfidence problem seen in modern neural networks when faced with unknown inputs~\cite{du2022vos, guo2017calibration, devries2018learning, piroli2023ls} and show the importance of counterbalancing methods like out-of-distribution detections and open-word classification.  
\section{Conclusion}
In this paper, we present the SemanticSpray++ dataset, which extends the SemanticSpray dataset~\cite{piroli2023SemanticSpray} with object labels for the camera and LiDAR data and semantic labels for the radar targets. 
We provide details on the annotation process and the challenges associated with it.  
Afterward, we present statistics for labels across the different sensor modalities and tasks.
Additionally, we evaluate the performance of popular 2D and 3D object detectors and 3D point cloud semantic segmentation methods and give insights on how spray affects their performances. 

In future work, we aim to provide additional labels for the dataset, like semantic masks for the camera image, and more proprieties on the object labels like occlusion levels.

\section{Acknowledgement}
We would like to thank the authors for the RoadSpray dataset~\cite{linnhoff2022roadspray}, on which our work is based.
Their effort in meticulously and comprehensively recording the large amount of data, allowed us to build the proposed dataset.

\bibliographystyle{IEEEtran}
\bibliography{mybib}

\end{document}